\documentclass[conference]{IEEEtran}
\IEEEoverridecommandlockouts

\usepackage{decar-common}
\usepackage{decar-dynamics}
\usepackage{decar-lie}
\usepackage{decar-post}

\usepackage{subcaption}
\usepackage[normalem]{ulem}
\usepackage{makecell}
\usepackage{eso-pic}

\newsavebox{\subfigbox}

\makeatletter
\AtBeginDocument{
  \check@mathfonts
}
\makeatother

\title{Range-Aided SLAM Initialization Exploiting Accurate Heading Information}

\author{
\IEEEauthorblockN{Isabel Lougheed}
\IEEEauthorblockA{
\textit{Department of Mechanical Engineering}\\
\textit{McGill University}\\
Montreal, QC, Canada\\
isabel.lougheed@mail.mcgill.ca}
\and
\IEEEauthorblockN{James Richard Forbes}
\IEEEauthorblockA{
\textit{Department of Mechanical Engineering}\\
\textit{McGill University}\\
Montreal, QC, Canada\\
james.richard.forbes@mcgill.ca}
\thanks{This work was supported by an Undergraduate Student Research Award (USRA) through the Natural Sciences and Engineering Research Council of Canada (NSERC), and Voyis Imaging Inc.}
}

\begin{document}

\AddToShipoutPictureFG*{%
  \AtPageLowerLeft{%
    \put(\LenToUnit{0.5\paperwidth},\LenToUnit{5mm}){%
      \makebox(0,0)[b]{%
        \parbox{0.92\paperwidth}{%
          \centering\footnotesize
          © 2026 IEEE. Personal use of this material is permitted.
          Permission from IEEE must be obtained for all other uses, in any
          current or future media, including reprinting/republishing this
          material for advertising or promotional purposes, creating new
          collective works, for resale or redistribution to servers or lists,
          or reuse of any copyrighted component of this work in other works.
        }%
      }%
    }%
  }%
}

\maketitle

\fontdimen16\textfont2=\fontdimen17\textfont2
\fontdimen13\textfont2=5pt
\begin{abstract}

This paper presents a novel initialization method for range-aided simultaneous localization and mapping (RA-SLAM). The general SLAM problem has a well-known separable structure where landmark and robot positions can be solved for in a linear fashion given known robot headings. This paper considers the case where highly accurate heading information is available, which is typical of autonomous underwater vehicle (AUV) navigation, to solve for the range transponder and robot positions. 
The proposed approach consists of two steps. 
First, using a generalized trust region subproblem (GTRS), the positions of the transponders relative to the AUV are solved for given the nonlinear range measurements. Second, the relative transponder positions and the known heading of the AUV are used to estimate the transponder and AUV positions by solving a linear least-squares problem. These transponder and AUV position estimates, combined with the highly accurate heading information, provide a reliable initialization method for the general nonlinear RA-SLAM problem. The effectiveness of the proposed approach is tested on a real-world AUV dataset where long baseline (LBL) range measurements are provided in concert with highly accurate heading information provided by an inertial navigation system (INS). 
\end{abstract}

\section{Introduction}
Autonomous underwater vehicles (AUVs) require knowledge of their position and orientation relative to their surroundings. 
AUVs commonly rely on long baseline (LBL) acoustic measurements for positioning, where LBL transponders act as fixed landmarks, and range measurements to these landmarks are measured via round-trip travel times \cite{farrell_aided_2008}. However, before AUV positions can be computed from LBL range measurements, the location of the LBL transponders must be solved for. One way to solve for the LBL transponder positions is to solve a range-aided simultaneous localization and mapping (RA-SLAM) problem. 

RA-SLAM uses range measurements to solve for landmark positions and robot poses.  
RA-SLAM is nonlinear and is often posed as a nonlinear least-squares problem, and then solved using the Gauss-Newton (GN) or Levenberg-Marquardt (LM) algorithms. 
R
To ensure GN or LM converges to an acceptable solution, a reasonable initial estimate of the landmark positions and robot poses is required.
Acquiring an initial estimate of the landmark locations can be difficult. For example, estimating LBL transponder locations is challenging when transponders are spaced far apart on the seafloor \cite{zhang_positioning_2018}. 
An initial estimate of the robot poses can be found through dead-reckoning odometry measurements.  However, drift in the dead-reckoned solution will accumulate over long distances, such as when an AUV is performing an underwater survey over a large area.

Existing RA-SLAM initialization methods generate an initial estimate of the robot position, the robot orientation, and landmark locations from a coupled landmark-pose structure \cite{delama_uvio_2023, delama_real-time_2025, papalia_score_2022, agarwal_rfm-slam_2016, carlone_fast_2014}. However, AUVs typically have access to highly accurate orientation information, or heading information in the planar case. For example, the Sonardyne SPRINT-Nav systems provide very accurate heading estimates, less than $0.05^\circ$ RMSE or $0.01^\circ$ RMSE, depending on the SPRINT-Nav model \cite{houlder2024_sonardyne_blog}.
As such, at least for initialization, it is unnecessary to solve for AUV headings. Rather, focus should be placed on solving for initial landmark position estimates, which in this paper are the LBL transponder locations, and initial AUV position estimates, to be used for RA-SLAM initialization. 

The general SLAM problem has a well-known separable structure where, given known robot headings, landmark and robot positions can be solved for using linear least-squares~\cite{khosoussi_exploiting_2015}.
This paper exploits this separable structure to solve for the initial LBL transponder locations and initial AUV positions in the plane, given highly accurate AUV heading information provided by an inertial navigation system (INS). 
To do so, the landmark position measurements cannot be in the form of range measurements, but must be in the form of relative position measurements between the LBL transponders and the AUV-mounted LBL transceiver, resolved in the robot body frame. This paper demonstrates how to express the LBL range measurements as relative landmark position measurements by solving a generalized trust region subproblem (GTRS).

The remainder of this paper is as follows. Related work is discussed in Section~\ref{sec:related_work}, and notation is reviewed in Section~\ref{sec:prelims}. The main solution methodology is presented in Section~\ref{sec:main_approach}. Section~\ref{sec:results} presents simulation and experimental results,
Section~\ref{sec:sensitivity_study} presents a sensitivity study, and the paper is drawn to a close in Section~\ref{sec:conclusion}.

\section{Related Work}
\label{sec:related_work}

Some notable RA-SLAM initialization methods include spectral graph partitioning \cite{olson_robust_2006} and spectral decomposition \cite{pmlr-v28-boots13}. However, these methods do not account for sensor noise, nor consider exploiting highly accurate heading information. 

The problem of estimating a source location, such as an LBL transponder or a UWB anchor, using range measurements has received significant attention.  The squared-range-based nonlinear least-squares (SR-LS) approach to solving this problem is nonconvex and is not optimal in the maximum likelihood sense due to the structure of the covariance matrix \cite{beck_exact_2008, cheung_least_2004}.  The non-convexity of the SR-LS problem has been addressed by reformulating the SR-LS as a GTRS \cite{beck_exact_2008, beck_solution_2012}.  However, these methods do not characterize uncertainty on the estimate.  Synthetic LBL is a similar RA-SLAM initialization method specific to LBL transponder initialization \cite{larsen_synthetic_2000, larsen_high_2000, wu_doppler_2018} where an AUV travels to points around a single LBL transponder measuring LBL range measurements. 
These methods rely on an accurate dead-reckoning solution to solve for the LBL transponder's position, but do not refine the robot's position.
The accuracy of these methods also depends significantly on the trajectory of the AUV.

Linear least-squares methods to initialize ultra-wideband (UWB) anchors exist, but often suffer from ill-posedness issues \cite{hausman_self-calibrating_2016}.
There are UWB initialization methods that initialize UWB anchors through nonlinear optimization. For example, UVIO acquires a coarse initial estimate using an optimal double method, and then refines the estimate through a nonlinear optimization \cite{delama_uvio_2023}.  A similar approach extends UVIO to realize a real-time initialization method for UWB anchors \cite{delama_real-time_2025}.  Both \cite{delama_uvio_2023,delama_real-time_2025} optimize keypoint selection based on the geometric dilution of precision (GDOP) to address observability issues associated with range measurements.  However, these methods estimate robot orientation, and do not consider the use of highly accurate heading information. These methods have also been developed for unmanned aerial vehicles (UAVs) operating over much smaller trajectories than AUVs typically do, therefore they may encounter scalability issues when applied to AUVs.

SCORE is an initialization method tested on AUVs where the nonconvex RA-SLAM problem is relaxed to a convex second-order conic problem (SOCP) \cite{papalia_score_2022}.  Similarly, CORA is a full RA-SLAM solution that uses a quadratically-constrained quadratic programming (QCQP) approach to RA-SLAM to relax the problem to a semidefinite program (SDP), where the problem is either randomly initialized or initialized through odometry measurements \cite{papalia_certifiably_2024}. However, a better initialization still helps with convergence. Additionally, SCORE and CORA do not take advantage of the highly accurate heading measurements that are available in underwater navigation.  These methods also struggle when the robot trajectory is not well constrained.  When there are not enough range measurements, SCORE tends to produce poor initializations, and CORA tends to loosen the SDP relaxation \cite{papalia_score_2022, papalia_certifiably_2024}.

Some notable methods that exploit the separable structure of SLAM include \cite{agarwal_rfm-slam_2016, carlone_fast_2014, carlone_angular_2012}.  RFM-SLAM estimates robot orientation through nonlinear optimization, and then estimates robot and landmark positions by solving a linear least-squares problem with relative feature measurements \cite{agarwal_rfm-slam_2016}.  However, this method is only tested on simulated data with many range measurements to many different landmarks. LAGO is a linear approximation for pose graph estimation that exploits the separable structure of SLAM to approximate the maximum likelihood solution \cite{carlone_fast_2014}.  However, LAGO is not robust when the pose graph has poor connectivity, for instance if the range measurements are sparse.  Additionally, MOLE2D \cite{carlone_angular_2012} only addresses orientation estimation.

Relative to existing literature, the novel contributions of this paper specific to RA-SLAM initialization for underwater navigation involving LBL transponders and highly accurate heading information are as follows. 
\begin{itemize}

\item Leveraging the separable structure of the SLAM problem when accurate heading information is available to estimate LBL transponder and AUV positions resolved in the world frame via linear least-squares, providing an initialization for nonlinear RA-SLAM. 

\item Converting LBL range measurements into relative positions between the LBL transponder and AUV-mounted transceiver resolved in the AUV body frame using a GTRS \cite{beck_exact_2008}.  The uncertainty associated with the LBL relative position estimate is characterized. These relative positions are then used in the aforementioned linear least-squares problem.

\item Experimentally validating the proposed approach against SCORE \cite{papalia_score_2022}, using a real-world AUV dataset containing highly accurate INS heading, Doppler velocity log (DVL)-derived velocity, LBL range, as well as groundtruth data.

\end{itemize}

The open-source implementation is available at 
\url{https://github.com/decargroup/RA_SLAM_initialization}.

\section{Notation}
\label{sec:prelims}

Column matrices are denoted $\mbf{r} \in \mathbb{R}^n$ and square or rectangular matrices are denoted $\mbf{A} \in \mathbb{R}^{m \times n}$. The $n \times n$ identity matrix is denoted $\mbf{1}$. The standard vector $2$ norm is denoted $\norm{\mbf{r}} = \sqrt{\mbf{r}^\trans \mbf{r}}$. A planar reference frame $\rframe{a}$ is composed of two orthonormal, dextral, basis vectors. The position of point $z$ relative to point $w$ resolved in $\rframe{a}$ is denoted $\mbf{r}^{zw}_a \in \mathbb{R}^2$, and when resolved in $\rframe{b}$ is denoted $\mbf{r}^{zw}_b \in \mathbb{R}^2$.  The direction cosine matrix (DCM) $\mbf{C}_{ab} \in SO(2)$ relates $\mbf{r}^{zw}_a$ and $\mbf{r}^{zw}_b$ via $\mbf{r}^{zw}_a = \mbf{C}_{ab} \mbf{r}^{zw}_b$ where  $\mbf{C}_{ab}^\trans = \mbf{C}_{ba}$ and $\mbf{C} \in SO(2)=\left\{\mbf{C} \in \mathbb{R}^{2 \times 2} \mid \mbf{C} \mbf{C}^{\trans} = \mbf{1} \right., \left.\operatorname{det} \mbf{C}=+1\right\}$ \cite{barfoot_state_2017}.  Time-varying quantities are indicated by the subscript $k$. For example, when point $z$ moves relative to point $w$ and $\mc{F}_b$ changes orientation relative to $\mc{F}_a$, $\mbf{r}^{z_k w}_{b_k}$ describes the position of point $z$ relative to point $w$ resolved in $\mc{F}_b$ at time $t_k$. Further, $\mbf{v}_{b_k}^{z_{k} w / a}$ denotes the velocity of point $z$ relative to point $w$ with respect to $\mc{F}_a$ resolved in frame $\mc{F}_b$ at time $t_k$. Using $\mbf{C}_{a b_k}$, $\mbf{r}_a^{z_k w} = \mbf{C}_{a b_k} \mbf{r}_{b_k}^{z_k w}$ and $\mbf{v}_a^{z_k w / a} = \mbf{C}_{a b_k} \mbf{v}_{b_k}^{z_k w / a}$. In this paper, point $w$ is the world datum, point $z_k$ is fixed to the AUV and can be assumed to be LBL transceiver position, and point $p$ is a landmark point corresponding to an LBL transponder. Furthermore, $\rframe{a}$ is the world frame and $\rframe{b}$ is the AUV's body frame. Often the AUV body frame at time $t_k$ is denoted $\mc{F}_{b_k}$. 

An uncertain or noisy DCM is given by 
\begin{align}
    \tilde{\mbf{C}}_{a b_{k}} = \mbf{C}_{a b_{k}} \exp(-{\theta}^\cross _{k}), \qquad \theta_k \sim \mc{N}(0,(\sigma^{\theta_{k}})^2) , \label{eq:heading_uncertainty}
\end{align}
where $\theta_k$ is the heading noise.  
Given that $\theta_k$ is small,
\begin{align}
    \exp(-{\theta}^\cross _{k}) \approx \mbf{1} - {\theta}^\cross _{k}, \quad
    {\theta}^\cross_{k} = \theta_{k} \mbs{\Gamma}, \quad  \mbs{\Gamma} = \begin{bmatrix}
        0 & -1 \\
        1 & 0
    \end{bmatrix}.
\end{align}

\section{Proposed Approach}
\label{sec:main_approach}

\subsection{Converting Range Measurements into Landmark Relative Position Measurements}
\label{sec:convert_range_meas_to_linear_meas}

\begin{figure}[h!]
    \centering
    \includegraphics[width = 0.36\textwidth]{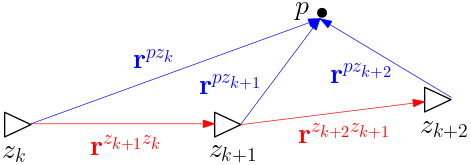}
    \caption{A robot observing a landmark at three different times, with landmark relative position measurements in blue and odometry measurements in red.}
    \label{fig:range_rel_posn_drawing}
\end{figure}

This section describes how to convert range measurements between a single LBL transponder and the robot-mounted LBL transceiver into relative position measurements between the LBL transponder and the robot-mounted LBL transceiver by solving a GTRS. 

Let $\mbf{r}_c^{z_k g}$ denote the position of point $z_k$ on the AUV relative to an arbitrary point $g$ resolved in an arbitrary frame $\mc{F}_c$, let $\mbf{r}_c^{p g}$ denote the position of a single LBL transponder referred to as the landmark point $p$ relative to point $g$ resolved $\mc{F}_c$. Let $\mbf{r}_{b_k}^{p z_k}$ denote the position of the landmark point $p$ relative to point $z_k$ on the AUV resolved in the AUV body frame $\mc{F}_{b_k}$ at time $t_k$, as seen in Figure~\ref{fig:range_rel_posn_drawing}. Point $z_k$ on the AUV can be assumed to be collocated with the LBL transceiver. An LBL range measurement is given by $r_k = \norm{\mbf{r}_c^{p z_k}} = \norm{\mbf{r}_c^{p g} - \mbf{r}_c^{z_k g}}$. The present goal is to use range measurements $r_k$, the known heading $\mbf{C}_{c b_k}$, and dead reckoned position estimates $\mbf{r}^{z_k g}_c$, to estimate $\mbf{r}^{p z_k}_{b_k}$. 

Consider the SR-LS problem
\begin{align}
    \min_{\mbf{r}_c^{p g}} \; \sum_{k=1}^{N} \left(\| \mbf{r}_c^{p g} - \mbf{r}_c^{z_k g} \|^2 + d^2 - r_k^2 \right)^2 , \label{srls}
\end{align}
where $d$ is used to take into account the height difference between the robot and the landmark when working in 2D, and $k = 1, 2, \ldots, N$ is a window of time. 
In \cite{beck_exact_2008} the SR-LS problem is transformed into a GTRS, 
\begin{align}
    \min_{\mbf{y}} \; \| \mbf{A} \mbf{y} - \mbf{b} \|^2 \mbox{ such that } \mbf{y}^{\trans} \mbf{D} \mbf{y} + 2 \mbf{f}^{\trans} \mbf{y} = 0  \label{eq:gtrs}
\end{align}
where
$
    \mbf{y} = 
    \begin{bmatrix}
        {\mbf{r}_c^{p g}}^{\trans} & \alpha
    \end{bmatrix}^{\trans}
$, $\alpha = \| \mbf{r}_c^{p g} \|^2$, 
$
    \mbf{D} = 
    \begin{bmatrix}
        \mbf{1} & \mbf{0} \\
        \mbf{0} & 0 \\
    \end{bmatrix}
$, 
$
    \mbf{f} = 
    \begin{bmatrix}
        \mbf{0} &
        -0.5
    \end{bmatrix}^\trans
$, and
\begin{align*}
    \mbf{A} = 
    \begin{bmatrix}
        -2 {\mbf{r}_c^{z_1 g}}^{\trans} & 1 \\
        \vdots & \vdots \\
        -2 {\mbf{r}_c^{z_N g}}^{\trans} & 1 \\
    \end{bmatrix}, \qquad \mbf{b} = 
    \begin{bmatrix}
        r_1^2 - \|\mbf{r}_c^{z_1 g}\|^2 - d^2\\
        \vdots \\
        r_N^2 - \|\mbf{r}_c^{z_N g}\|^2 - d^2\\
    \end{bmatrix} .
\end{align*}
Instead of using all range measurements to form the GTRS, optimal keypoints can be selected to minimize the position dilution of precision (PDOP).  The PDOP is
\begin{align}
    PDOP &= \sqrt{\mbf{G}_{11} + \mbf{G}_{22} + \mbf{G}_{33}},
\end{align}
where
\begin{align}
    \mbf{H} &= \begin{bmatrix}
        \frac{\mbf{r}_c^{p g} - \mbf{r}_c^{z_1 g}}{\| \mbf{r}_c^{p g} - \mbf{r}_c^{z_1 g} \|} & 1 \\
        \vdots & \vdots \\ 
        \frac{\mbf{r}_c^{p g} - \mbf{r}_c^{z_k g}}{\| \mbf{r}_c^{p g} - \mbf{r}_c^{z_k g} \|} & 1
    \end{bmatrix}, \qquad \mbf{G} = (\mbf{H}^\trans \mbf{H})^{-1}.
\end{align}
The PDOP is large when the rows of $\mbf{H}$ are nearly linearly dependent, indicating that the localization problem is poorly observable \cite{farrell_aided_2008}.  Optimal keypoints are selected using a greedy keypoint selection algorithm similar to \cite{cossette_relative_2021}.

After solving the GTRS for $\mbf{r}_c^{p g}$, $\mbf{r}_{b_k}^{p z_k}$ must be solved for at each time $t_k$ within the window $k = 1, 2, \ldots, N$.
Doing so amounts to computing 
\begin{align}
    \mbf{r}_{b_k}^{p z_k} = \mbf{C}_{c b_k}^{\trans} ( \mbf{r}_c^{p g} - \mbf{r}_c^{z_k g} ) . \label{rel_pos_conversion}
\end{align}
Equation~\eqref{rel_pos_conversion} elucidates why both point $g$ and reference frame $\mc{F}_c$ are arbitrary. As can be seen in \eqref{rel_pos_conversion}, the point $g$ cancels out when computing the relative position $\mbf{r}_c^{p z_k} = \mbf{r}_c^{p g} - \mbf{r}_c^{z_k g}$, and reference frame $\mc{F}_c$ cancels out when computing $\mbf{r}_{b_k}^{p z_k} = \mbf{C}_{c b_k}^{\trans} \mbf{r}_c^{p z_k}$. 
In this paper, $\mbf{r}_c^{z_k g}$ is assumed to be computable via dead-reckoning using DVL and INS data, which is quite accurate over short times, resulting in $w$ and $\mc{F}_a$ being natural choices for points $g$ and $\mc{F}_c$. However, in some situations it may be worthwhile to consider alternative choices for $g$ and $\mc{F}_c$.

Having solved for $\mbf{r}_{b_k}^{p z_k}$, the uncertainty must be properly characterized. 
Consider
\begin{align}
    \tilde{d} &= d + \zeta, &\zeta &\sim \mc{N}(0,(\sigma^{d})^2), \\
    \tilde{r}_k &= r_k + \eta_k, &\eta_k &\sim \mc{N}(0,(\sigma^{r_k})^2), \\
    \tilde{\mbf{v}}_{b_{k}}^{z_{k} w/a} &= \mbf{v}^{z_{k} w/a}_{b_{k}} + \mbs{\gamma}_{k}, &\mbs{\gamma}_{k} &\sim \mc{N}(0,\mbf{Q}_{k}), \\
    \tilde{\mbf{C}}_{a b_{k}} &= \mbf{C}_{a b_{k}} \exp(-{\psi}^\cross _{k}), &{\psi}_{k} &\sim \mc{N}(0,(\sigma^{\theta_{k}})^2), \\
    \tilde{\mbf{r}}^{z_k w}_{a} &= {\mbf{r}}^{z_k w}_{a} + \mbs{\xi}_k, &\mbs{\xi}_{k} &\sim \mc{N}(0,\mbf{R}_{dr}),
\end{align}
where $\zeta, \eta_k, \mbs{\gamma}_k, \psi_k$, and $\mbs{\xi}_k$ are the noise variables associated with the depth measurements, range measurements, DVL-based velocity measurements, heading measurements, and dead-reckoned trajectory estimate respectively.

Before computing the uncertainty associated with each $\mbf{r}_{b_k}^{p z_k}$, the uncertainty associated with the landmark position estimate $\mbf{r}^{p w}_{a}$ must be characterized.
Although a solution to a GTRS provides $\mbf{r}^{p w}_{a}$, the uncertainty associated with $\mbf{r}^{p w}_{a}$ can still be characterized assuming a standard nonlinear least-squares approach has been used to compute $\mbf{r}^{p w}_{a}$.  
The reason is that, ultimately, a solution to a nonlinear least-squares problem has been found, just by solving a GTRS rather than using GN, LM, or any other nonlinear least-squares solver. 
The covariance associated with the solution to the nonlinear least-squares problem is given by
\begin{align}
    \mbs{\Sigma} = \sum_{k = 1}^N \sigma_{e_k}^2(\mbf{J}_k^\trans \mbf{J}_k)^{-1}, \label{Sigma}
\end{align}
where $\mbf{J}_k = 2(\mbf{r}^{p w}_a - \mbf{r}^{z_k w}_a)^\trans$ is the Jacobian associated with range measurement $r_k$ and $\sigma_{e_k}^2$ is the residual variance \cite{barfoot_state_2017}.  The residual variance $\sigma_{e_k}^2$ can be found considering the noise within the terms appearing in ${e_k} = \|\mbf{r}^{p w}_a - \mbftilde{r}^{z_k w}_a\|^2 + \tilde{d}^2 - \tilde{r}_k^2$ that appears in \eqref{srls}, solving for the overall noise $w_e$, and then computing $\sigma_{e_k}^2 = E[w_e^2]$ where $E[\cdot]$ is the expectation operator. To this end, 
\begin{align}
    e_k 
    &= \|\mbf{r}^{p w}_a - (\mbf{r}^{z_k w}_a + \mbs{\xi}_k)\|^2 + (d + \zeta)^2 - (r_k + \eta_k)^2 , \notag \\
    w_e & = \underbrace{2\begin{bmatrix}
        (\mbf{r}^{z_k w}_a - \mbf{r}^{p w}_a)^{\trans} & d & -r_k
    \end{bmatrix}}_{\mbf{M}_k} \begin{bmatrix}
        \mbs{\xi}_k \\ \zeta \\ \eta_k
    \end{bmatrix}. \label{srls_noise}
\end{align}
Therefore, the residual covariance is
\begin{align}
    \small
    {\sigma}_{e_k}^2 = \mbf{M}_k
    \begin{bmatrix}
        \mbf{R}_{dr} & & \\
        & (\sigma^d)^2 & \\
        & & (\sigma^{r_k})^2
    \end{bmatrix} \mbf{M}_k^\trans,
\end{align}
and the uncertainty associated with the landmark estimate $\mbf{r}^{p w}_{a}$ can now be defined as $\mbs{\epsilon} \sim \mc{N}(0,\mbs{\Sigma})$, where $\mbs{\Sigma}$ comes from \eqref{Sigma}.  

Next, the uncertainty associated with the relative position estimate $\mbf{r}^{p z_k}_{b_k}$ can now be found by expanding \eqref{rel_pos_conversion} to include noise $\mbs{\nu}_{k}$, solving for $\mbs{\nu}_{k}$, and then computing $\mbf{R}_k = E[\mbs{\nu}_{k} \mbs{\nu}_{k}^\trans]$. To this end,
\begin{align}
    & \mbf{r}^{p z_k}_{b_k} + \mbs{\nu}_{k} \notag \\
    & =
    (\mbf{C}_{ab_k} \exp (-\psi_k^\cross))^{\trans} ((\mbf{r}^{p w}_{a} + \mbs{\epsilon}) - (\mbf{r}^{z_k w}_{a} + \mbs{\xi}_k)) \notag \\
    & =
    \underbrace{\mbf{C}_{ab_k}^{\trans} (\mbf{r}^{p w}_{a}  - \mbf{r}^{z_k w}_{a})}_{\mbf{r}_{b_k}^{p z_k}} \notag \\
    & +
    \underbrace{\begin{bmatrix}
        \mbf{C}_{ab_k}^{\trans} & -\mbf{C}_{ab_k}^{\trans} & \mbs{\Gamma} \mbf{C}_{ab_k}^{\trans} (\mbf{r}^{p w}_{a}  - \mbf{r}^{z_k w}_{a})
    \end{bmatrix}
    \begin{bmatrix}
        \mbs{\epsilon} \\ \mbs{\xi}_k \\ \psi_k
    \end{bmatrix}}_{\mbs{\nu}_{k}}. 
    \label{eq:beacon_rel_robot_in_robot_F_noise}
\end{align}
The uncertainty associated with the dead-reckoned robot position estimate $\mbf{r}^{z_k w}_a$, that being $\mbs{\xi}_k$, can be computed via
\begin{align}
     \mbf{r}^{z_k w}_a + \mbs{\xi}_k \notag 
    & = \mbf{r}^{z_{k-1} w}_a + \Delta t \mbf{C}_{ab_{k-1}} \mbf{v}_{b_{k-1}}^{z_{k-1} w/a} \notag \\
    & + 
    \underbrace{\begin{bmatrix}
        -\Delta t \mbf{C}_{ab_0} \mbs{\Gamma} \mbf{v}_{b_{0}}^{z_{0} w/a} & \hdots & \Delta t \mbf{C}_{ab_1} & \hdots
    \end{bmatrix} 
    \begin{bmatrix}
         {\psi}_{1} \\ \vdots \\ \mbs{\gamma}_{0} \\ \vdots
     \end{bmatrix}}_{\mbs{\xi}_k}.
     \label{dr_noise}
\end{align}
Now $\mbs{\xi}_k$ can be substituted into the expression for $\mbs{\nu}_{k}$ in \eqref{eq:beacon_rel_robot_in_robot_F_noise}. Computation of $\mbf{R}_k = E[\mbs{\nu}_{k} \mbs{\nu}_{k}^\trans]$ follows in a tedious manner. 

With the estimates of landmark positions resolved in the robot's body frame along with their respective covariances, these estimates can be treated as linear relative positions in a SLAM problem that solves for landmark locations and robot positions. 

\subsection{SLAM with Known Heading as a Linear Least-Squares Problem}

With the range measurements transformed into linear relative position measurements, and accurate heading information available, the separable property of SLAM can be exploited to estimate the landmark and robot positions at all times in a batch fashion using linear least-squares. For simplicity of exposition, uncertainty in all quantities will be neglected.

Consider the process model
\begin{align}
    \mbf{r}^{z_k w}_{a} & = \mbf{r}^{z_{k-1} w}_{a} + \Delta t  \mbf{C}_{a b_{k-1}} \mbf{v}_{b_{k-1}}^{z_{k-1} w/a}, \label{eq:process_model_1} 
\end{align}
and the measurement model
\begin{align}
    \mbf{y}_k^i & = \mbf{r}^{p_i z_k}_{b_k} = \mbf{C}_{a b_k}^\trans ( \mbf{r}^{p_i w}_a - \mbf{r}^{z_k w}_a ) , \notag \\
    \mbf{C}_{a b_k} \mbf{y}_k^i & = \mbf{r}^{p_i w}_a - \mbf{r}^{z_k w}_a ,
    \label{eq:meas_model_batch}
\end{align}
where, critically, $\mbf{C}_{a b_{k-1}}$ and $\mbf{C}_{a b_k}$ are assumed to be known. The unknowns to be solved for are the $i = 1, 2, \ldots, L$ landmark positions $\mbf{r}^{p_i w}_a$ appearing in \eqref{eq:meas_model_batch}, and the $k = 1, 2, \ldots, K$ robot positions $\mbf{r}^{z_k w}_a$. 
Together \eqref{eq:process_model_1} and \eqref{eq:meas_model_batch} form the basis for a linear-least squares problem, that being
\begin{align}
    \bbm
        -\mbf{1} & \mbf{1} & \mbf{0} \\
        \mbf{0} & -\mbf{1} & \mbf{1}
    \ebm
    \bbm
        \mbf{r}_a^{z_{k-1} w} \\
        \mbf{r}_a^{z_k w} \\
        \mbf{r}^{p_i w}_a
    \ebm
    =
    \bbm
        \Delta t \mbf{C}_{a b_{k-1}} \mbf{v}_{b_{k-1}}^{z_{k-1} w/a} \\
        \mbf{C}_{a b_{k-1}} \mbf{y}_k^i
    \ebm . \label{eq:based_Ax=b_prob}
\end{align}

As a concrete example, consider estimating three robot positions, $\mbf{r}_a^{z_1 w}, \mbf{r}_a^{z_2 w}, \mbf{r}_a^{z_3 w}$, and one landmark position, $\mbf{r}_a^{p w}$. Specifying $\mbf{r}_a^{z_0 w}$ to render the SLAM problem solvable, solving for the unknown robot positions and single landmark amounts to solving
\begin{align}
    \begin{bmatrix}
    \mbf{1} & \mbf{0} &  \mbf{0} & \mbf{0} \\
    -\mbf{1} & \mbf{1} &  \mbf{0} & \mbf{0} \\
    \mbf{0} & -\mbf{1} & \mbf{1} &  \mbf{0} \\
    \hdotsfor{4} \\
    \mbf{0} &  \mbf{0} & \mbf{0} & \mbf{1} \\
    -\mbf{1} &  \mbf{0} & \mbf{0} & \mbf{1} \\
    \mbf{0} &  -\mbf{1} & \mbf{0} & \mbf{1} \\
    \mbf{0} &  \mbf{0} & -\mbf{1} & \mbf{1} \\
\end{bmatrix}
\begin{bmatrix}
    \mbf{r}^{z_1 w}_a \\ \mbf{r}^{z_2 w}_a \\ \mbf{r}^{z_3 w}_a \\
    \mbf{r}^{p w}_a 
\end{bmatrix}
=
\begin{bmatrix}
    \mbf{r}^{z_0 w}_a + \Delta t \mbf{C}_{a b_0} \mbf{v}^{z_0 w / a}_{b_0} \\
    \Delta t \mbf{C}_{a b_1} \mbf{v}^{z_1 w / a}_{b_1} \\
    \Delta t \mbf{C}_{a b_2} \mbf{v}^{z_2 w / a}_{b_2} \\
    \hdotsfor{1} \\
    \mbf{C}_{a b_0} \mbf{y}_0 + \mbf{r}^{z_0 w}_a \\ 
    \mbf{C}_{a b_1} \mbf{y}_1 \\
    \mbf{C}_{a b_2} \mbf{y}_2 \\
    \mbf{C}_{a b_3} \mbf{y}_3
\end{bmatrix}, \notag
\end{align}
which has a natural $\mbf{A} \mbf{x} = \mbf{b}$ structure that can be solved as a linear least-squares problem. Notice all ``knowns" or measurements appear in the $\mbf{b}$ matrix. 

Returning to the incorporation of uncertainty, recall from Section~\ref{sec:convert_range_meas_to_linear_meas} that the noise on $\mbf{y}_k^i = \mbf{r}_{b_k}^{p_i z_k}$ depends on the noise on the velocity $\mbf{v}_{b_{k-1}}^{z_{k-1} w / a}$. However, the process model \eqref{eq:process_model_1} depends on the velocity $\mbf{v}_{b_{k-1}}^{z_{k-1} w / a}$ as well. As such, the process model \eqref{eq:process_model_1} and measurement model \eqref{eq:meas_model_batch} are correlated. This correlation should be accounted for when formulating an overall covariance matrix to be used when solving the least-squares problem in a weighted fashion. However, to avoid the complexities of a non-block-diagonal covariance matrix, this correlation is neglected.

\section{Simulations and Experiments}
\label{sec:results}

\subsection{Simulated Data}
\label{sec: sim_data}
The proposed method was tested on a simulated dataset where 
LBL range measurements are consistently available at $1~(\si{Hz})$
with a standard deviation of $0.32~(\si{m})$.
DVL-derived velocity measurements and heading measurements both have a measurement frequency of $10~(\si{Hz})$ and standard deviations of $0.32~(\si{m/s})$ and $0.5^\circ$, respectively.

\begin{figure}
    \centering
    \includegraphics[width = 0.5\textwidth]{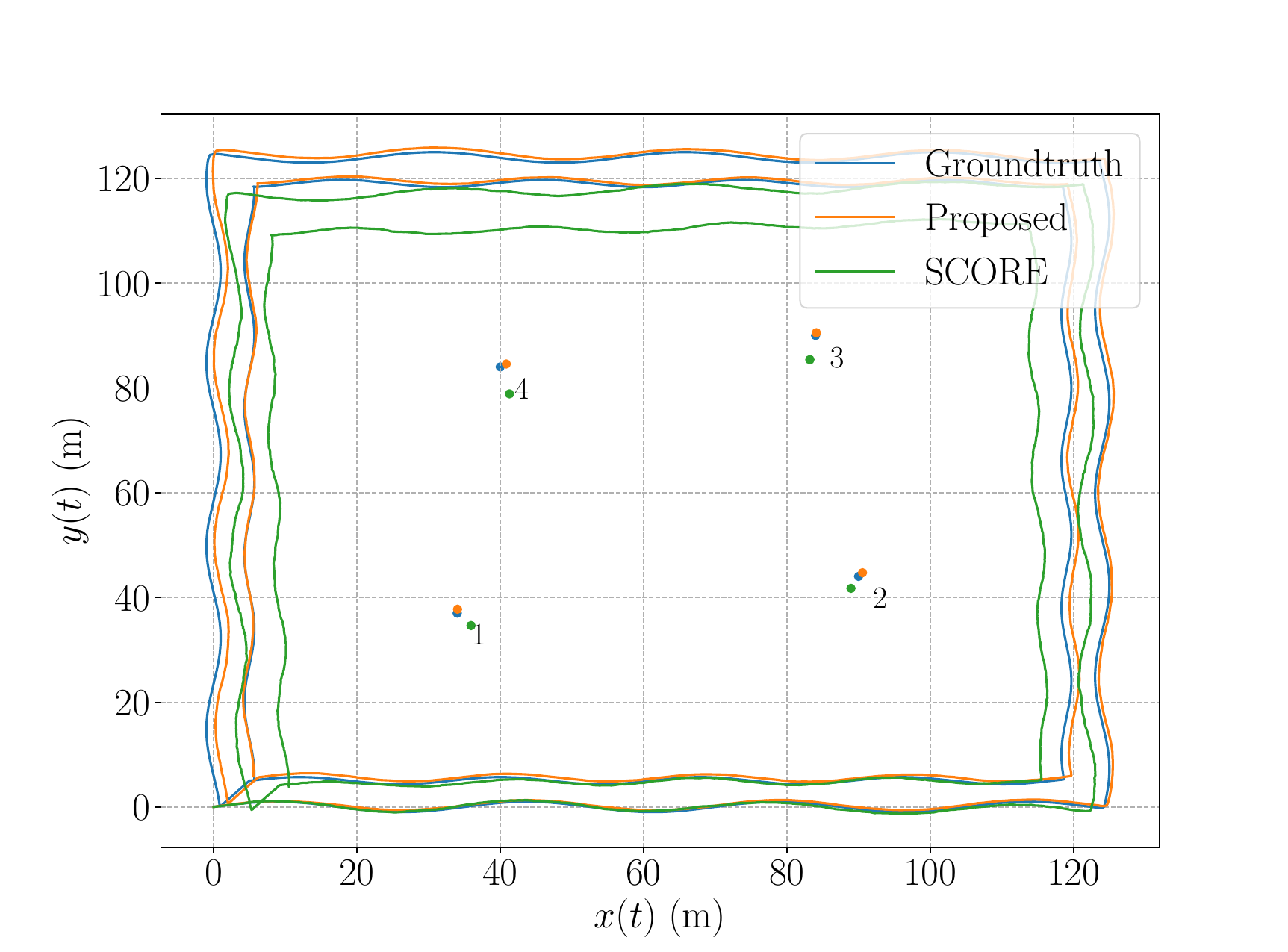}
    \caption{True vs. estimated landmark and AUV positions for a simulated dataset with consistent LBL range measurements.}
    \label{fig:true_vs_est_sim}
\end{figure}

Table~\ref{l_error_comparison_sim} presents the absolute position errors for all landmarks and the RMSE of the robot trajectory of the proposed method compared to SCORE \cite{papalia_score_2022}.  Figure~\ref{fig:true_vs_est_sim} presents the landmark and robot position estimates produced by the proposed method and SCORE compared to groundtruth.  As shown in both Table~\ref{l_error_comparison_sim} and Figure~\ref{fig:true_vs_est_sim}, the proposed method produces more accurate landmark and robot position estimates compared to SCORE.  Additionally, the proposed method produces an uncertainty associated with each state estimate, whereas SCORE does not.

\begin{table}
\begin{center}
    \begin{tabular}{|c|c|c|}
        \hline
        Metric & Proposed (\si{m}) & SCORE (\si{m}) \\
        \hline
        Landmark 1 error & \textbf{0.746} & 3.069 \\
        Landmark 2 error & \textbf{0.889} & 2.508 \\
        Landmark 3 error & \textbf{0.524} & 4.690 \\
        Landmark 4 error & \textbf{1.018} & 5.308 \\
        \hline
        Robot trajectory RMSE & \textbf{0.966} & 5.511 \\
        \hline
    \end{tabular}
    \caption{Comparison of absolute landmark position errors and robot trajectory RMSE for simulated dataset.}
    \label{l_error_comparison_sim}
\end{center}
\end{table}

\subsection{Experimental Data}
The proposed method was tested on an experimental AUV dataset collected by Sonardyne International Ltd in Plymouth, UK in March of 2024. The dataset contains heading measurements, depth measurements, DVL-derived velocity measurements, and LBL range measurements to two transponders on the seafloor, referred to as landmarks 2301 and 2801.  The heading uncertainty has a standard deviation of $8.7266 \times 10^{-3}~(\si{rad}) = 0.5^\circ$.  Groundtruth aiding was provided by both GPS and LBL.

\begin{figure}
    \centering
    \includegraphics[width = 0.5\textwidth]{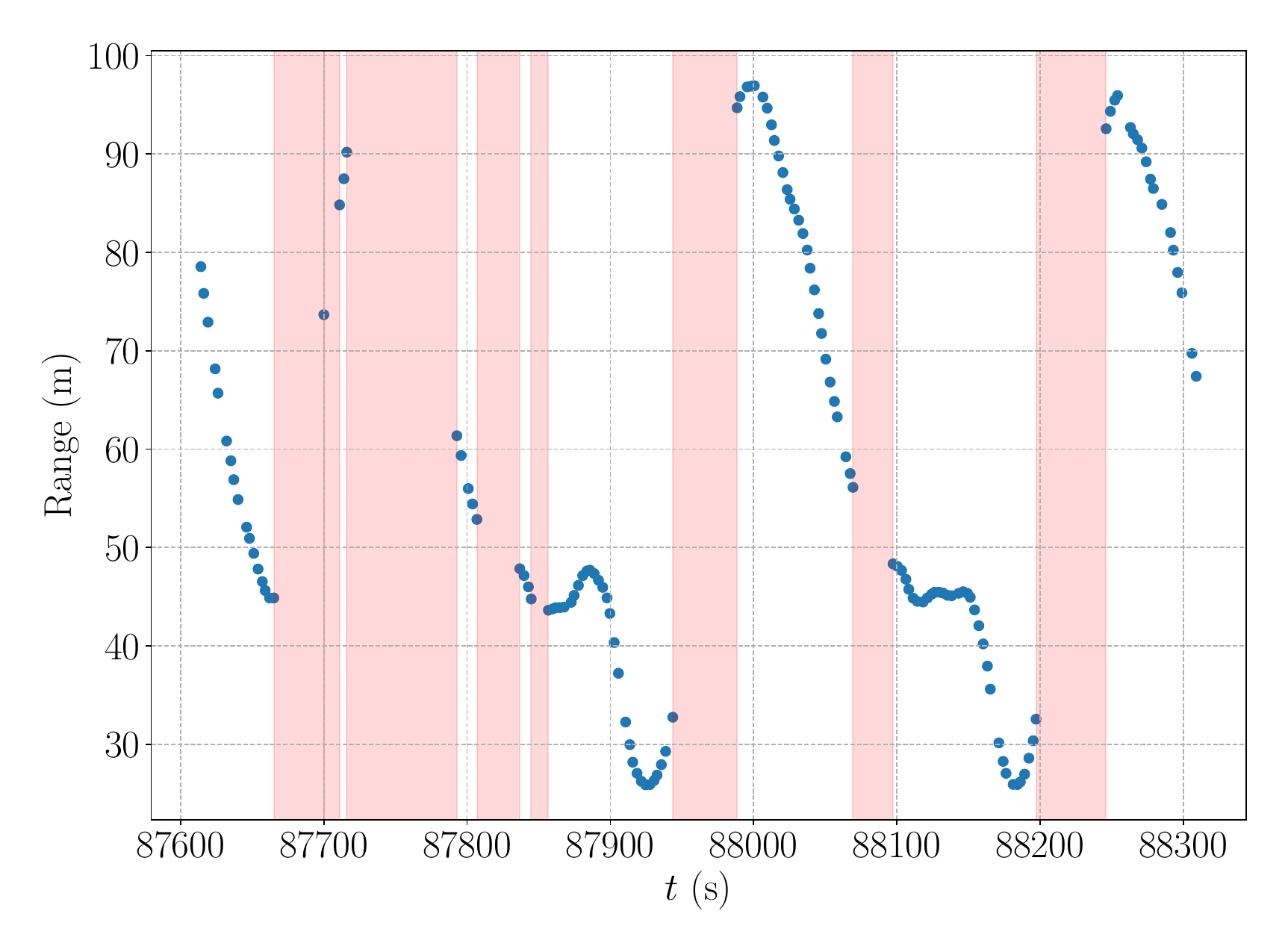}
    \caption{LBL measurements from landmark 2301. The range measurements are denoted in blue and the red horizontal lines indicate where the LBL measurements drop out.}
    \label{fig:ranges_2301}
\end{figure}

It is important to note that the LBL measurements are quite sparse, with a measurement frequency of approximately $0.36~(\si{Hz})$. Additionally, there are LBL measurement ``dropouts".  Notably, landmark 2301 is unavailable along one of the bends of the AUV trajectory, meaning that the associated LBL measurements drop out for extended periods of time, up to 77 seconds. This is shown in Figure~\ref{fig:ranges_2301}. This is a difficult dataset for RA-SLAM initialization and solution methods. 

\begin{table}
\begin{center}
    \begin{tabular}{|c|c|c|}
        \hline
        Metric & Proposed (\si{m}) & SCORE (\si{m}) \\
        \hline
        Landmark 2301 error & \textbf{1.735} & 21.248 \\
        Landmark 2801 error & \textbf{1.034} & 20.121 \\
        \hline
        Robot trajectory RMSE & \textbf{0.292} & 8.963 \\
        \hline
    \end{tabular}
    \caption{Comparison of absolute landmark position errors and robot trajectory RMSE for experimental dataset.}
    \label{l_error_comparison}
\end{center}
\end{table}

Table~\ref{l_error_comparison} presents the absolute position error of both landmarks after solving both the proposed method and SCORE.  The proposed method produces accurate estimates of both landmark and AUV positions, even with such a difficult dataset.  Notably, the landmark errors produced by the proposed method stay within the $\pm 3\sigma$ uncertainty bounds for both landmarks, indicating consistency.

\begin{figure}
    \centering
    \includegraphics[width = 0.5\textwidth]{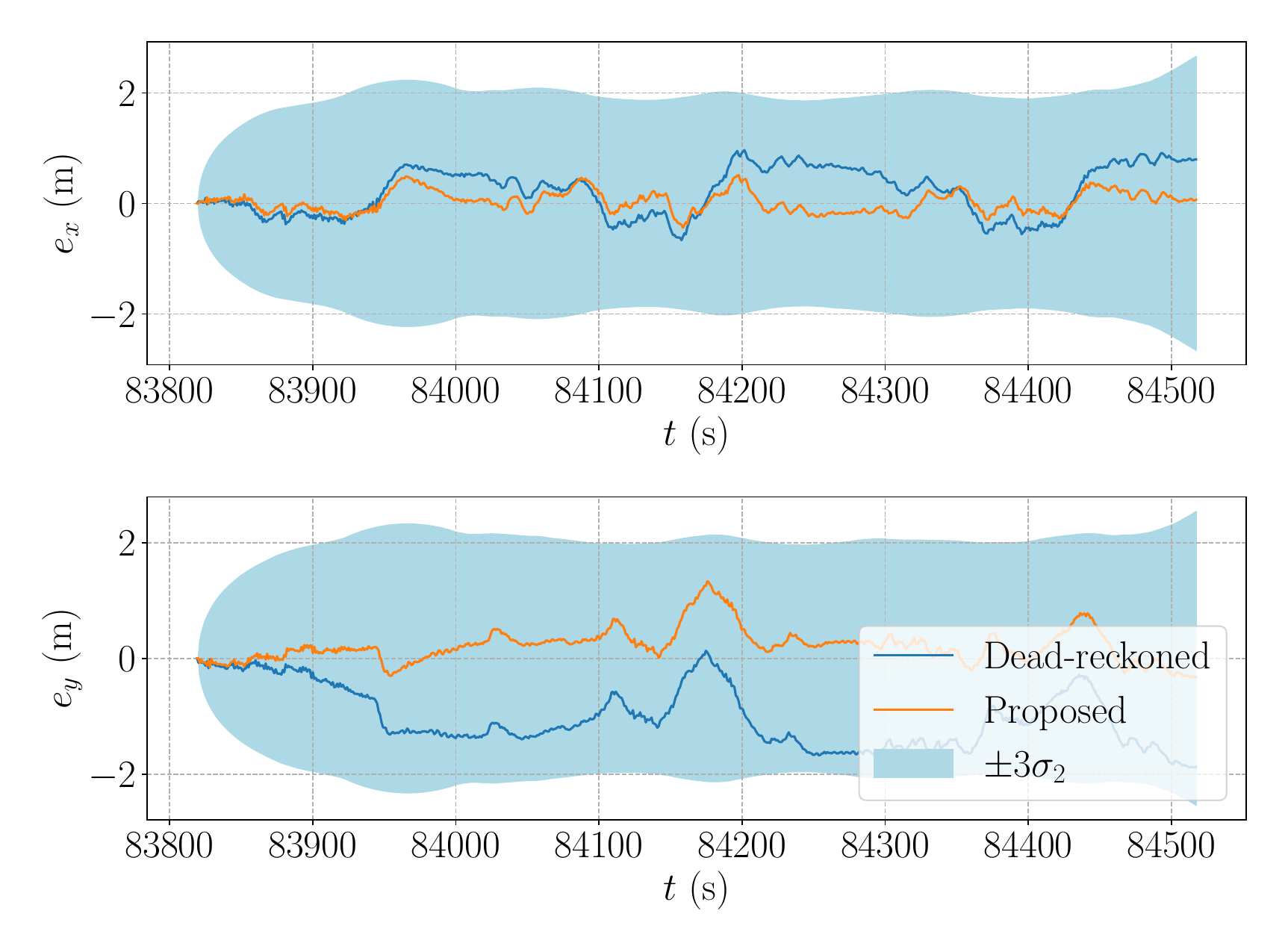}
    \caption{Robot position error compared to the position error from dead-reckoning. The $\pm 3 \sigma$ bounds are associated with the proposed method.}
    \label{fig:robot_error}
\end{figure}

As shown in Figure~\ref{fig:robot_error}, the proposed method produces a better estimate of AUV positions than the position estimate provided by dead-reckoning. Additionally, as indicated by the $\pm 3 \sigma$ bounds, the errors computed by the proposed method are consistent.

\begin{figure}
    \centering
    \includegraphics[width = 0.5\textwidth]{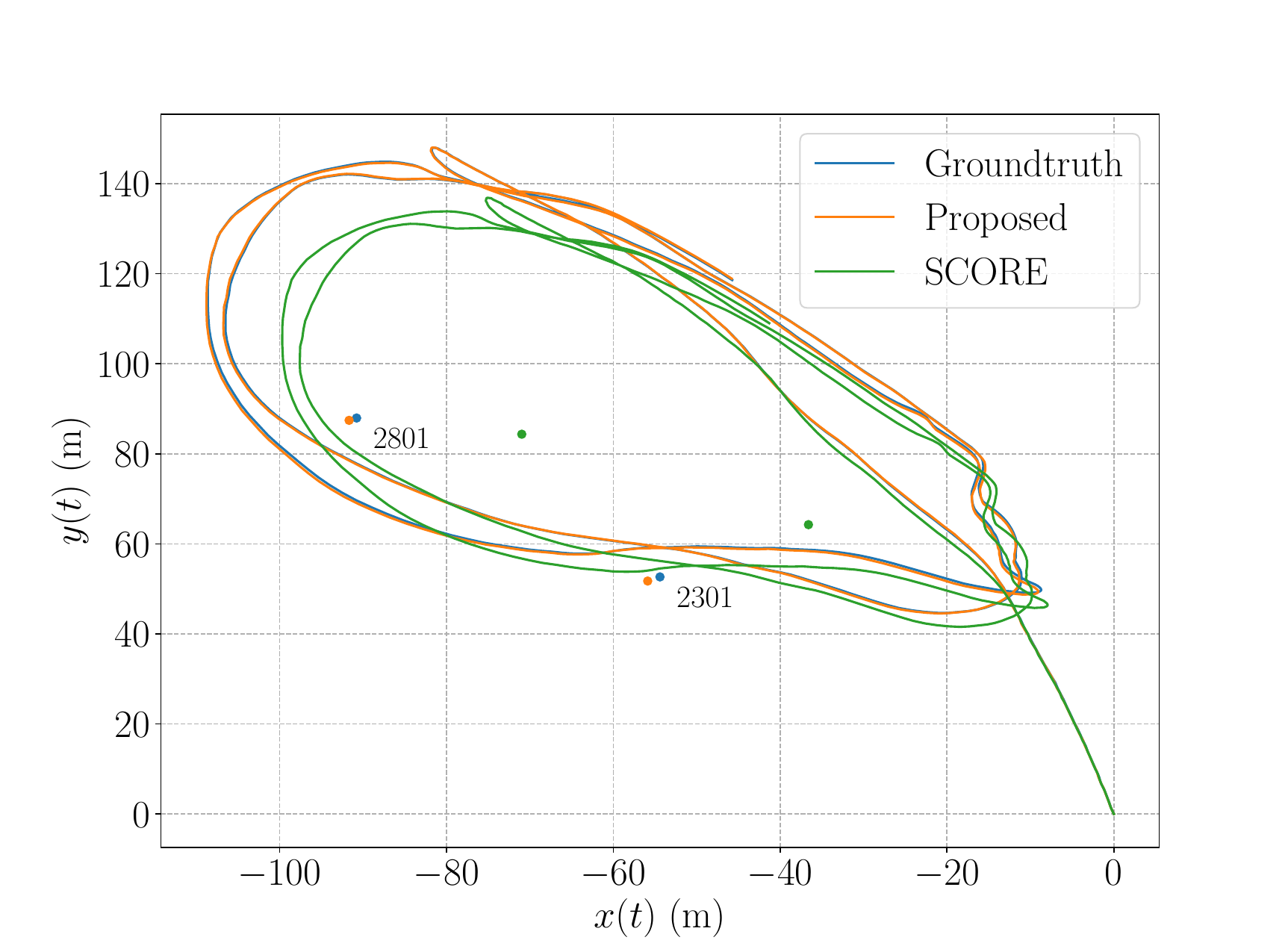}
    \caption{True vs. estimated landmark and AUV positions for an experimental dataset.}
    \label{fig:true_vs_est}
\end{figure}

Figure~\ref{fig:true_vs_est} compares the landmark and AUV position estimates produced by the proposed method and SCORE \cite{papalia_score_2022} compared to groundtruth.  SCORE produces poor landmark and position estimates relative to both the proposed method and groundtruth, highlighting the difficult nature of the dataset, and the capabilities of the proposed  method relative to SCORE. In this case, the proposed method would be a better choice for RA-SLAM initialization than SCORE. 

\section{Sensitivity Study}
\label{sec:sensitivity_study}
\begin{figure}
   \centering
   \includegraphics[width = 0.5\textwidth]{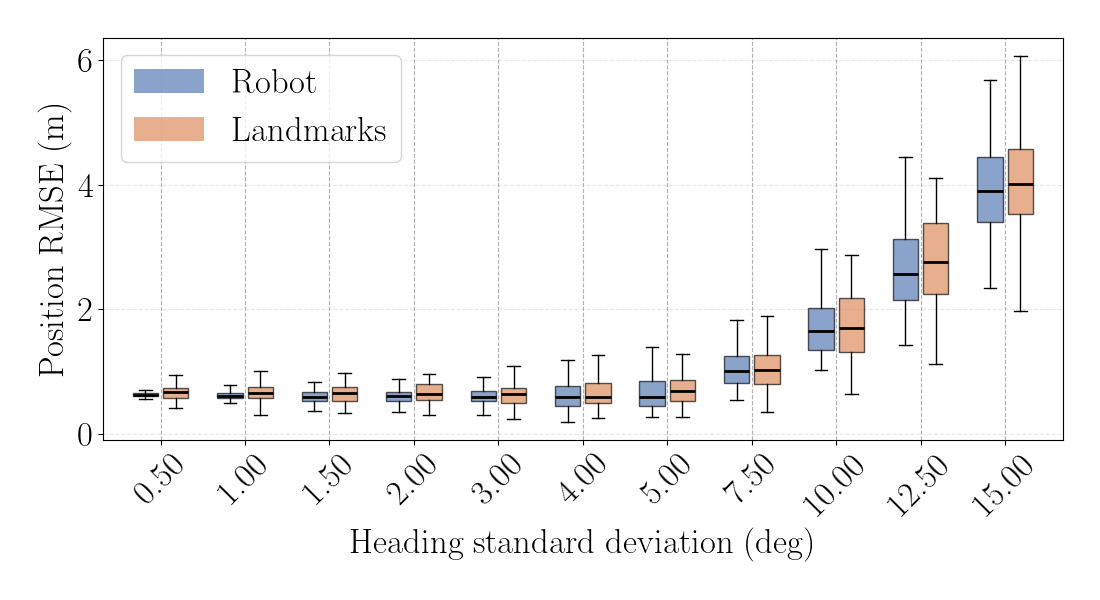}
   \caption{Robot and landmark position estimate RMSE for different levels of heading measurement noise, for 100 Monte Carlo trials.}
   \label{fig:ablation_rmse}
\end{figure}

To determine how accurate the heading measurements need to be for the proposed method to perform well, a sensitivity study is presented where the standard deviation of heading noise is varied.  This study is tested on the simulated dataset from Section~\ref{sec: sim_data}.  Figure~\ref{fig:ablation_rmse} shows how both the robot and landmark estimation errors are affected by different levels of heading measurement noise, where a significant performance degradation can be seen for heading standard deviations above $5.0^\circ$.  However, the proposed method shows promising results for heading standard deviations of $5.0^\circ$ and below, highlighting robustness to moderate levels of heading noise.

\section{Conclusion}
\label{sec:conclusion}

Given the nonlinear nature of the RA-SLAM problem, initialization is important to ensure convergence.  It can be difficult to acquire an initial estimate of landmark locations and robot poses, especially in the context of underwater navigation, where there are long trajectories over large areas.

This paper presents a novel initialization method for RA-SLAM that takes advantage of the separable property of SLAM. In particular, heading is not estimated when highly accurate heading information is available. Rather, just landmark and robot positions are estimated. To exploit the separable property of SLAM, range measurements are used to compute pseudo position measurements of the landmarks relative to the robot.

Focusing in on underwater navigation, the proposed approach is tested on a simulated dataset and a difficult real-world AUV dataset.
The proposed method produces accurate robot and landmark position estimates, even with sparse LBL measurements. The proposed approach performs admirably relative to SCORE.

\section*{Acknowledgment}

The authors thank Thomas Hitchcox for guidance, Sonardyne International Ltd. for the data, and Iman Shames for pointing the authors to the various works associated with the SR-LS problem \cite{beck_exact_2008, cheung_least_2004, beck_solution_2012}.



\printbibliography

\end{document}